\documentclass[letterpaper, 10 pt, journal, twoside]{ieeetran} 
\IEEEoverridecommandlockouts                               
\usepackage{etoolbox}
\makeatletter
\patchcmd{\@makecaption}
  {\scshape}
  {}
  {}
  {}
\makeatother

\IEEEoverridecommandlockouts                               

\usepackage{bm}

\usepackage{graphicx}                       
\usepackage{graphics}                       
\usepackage{epsfig}                         
\usepackage[tight,footnotesize]{subfigure}  
\graphicspath{./pics/RAL_fig_for_revise}
\graphicspath{./pics/RAL_new_fig}
\usepackage{amssymb,amsmath}
\usepackage{mdwmath}
\usepackage{commath}   
\usepackage{eqparbox}
\usepackage{mathtools}
\usepackage[utf8]{inputenc} 
\usepackage[english]{babel}
\usepackage{amsmath,amsfonts,amssymb}

\newcommand{\beq}{\begin{equation}}
\newcommand{\eeq}{\end{equation}}
\newcommand{\bear}{\begin{eqnarray}}
\newcommand{\bears}{\begin{eqnarray*}}
\newcommand{\eear}{\end{eqnarray}}
\newcommand{\eears}{\end{eqnarray*}}
\newcommand{\bdm}{\begin{displaymath}}
\newcommand{\edm}{\end{displaymath}}
\newcommand{\lba}{\left[\begin{array}}
\newcommand{\ear}{\end{array}\right]}

\newcommand{\degree}{^\circ}     
\usepackage{tabularray}

\usepackage{stfloats}                       
\usepackage{url} 
\usepackage{hyperref}
\usepackage{cite}                           
\usepackage[T1]{fontenc} 
\usepackage{tabularx}
\usepackage{diagbox} 
\usepackage{color}
\usepackage{multirow}
\usepackage[table]{xcolor}
\usepackage{longtable}
\usepackage{booktabs} 			
\usepackage{xcolor,colortbl}    

\usepackage{multicol}
\usepackage{graphicx}
\usepackage{placeins}
\usepackage{overpic}
\usepackage{amssymb}
\usepackage{pifont}

\usepackage[fleqn]{nccmath}

\title{NeuralTouch: Neural Descriptors for\\ Precise Sim-to-Real Tactile Robot Control}

\author{
Yijiong Lin*, 
Bowen Deng, 
Keju Pu,
Chenghua Lu,
Max Yang,
Efi Psomopoulou,
Nathan F. Lepora \\
\thanks{YL and NFL were supported by ARIA Robot
Dexterity award SMRB-PR01-P19. EP was supported by ARIA Robot
Dexterity award SMRB-PR01-P04/Artimus.

All authors are with the Department of Engineering Mathematics and Bristol Robotics Laboratory, University of Bristol, Bristol BS8 1UB, U.K. (*Corresponding author: yijiong.lin@bristol.ac.uk)}
}

\begin{document}
\maketitle
 
\begin{abstract}

Grasping accuracy is a critical prerequisite for precise object manipulation that may require careful alignment between the robot hand and an object. Neural Descriptor Fields (NDFs) offer a promising vision-based method to generate grasping poses that generalize across object categories. However, NDFs alone can produce inaccurate poses due to imperfect camera calibration, incomplete point clouds, and object variability. Alternatively, tactile sensing enables more precise contact, but existing approaches limit policies to simple predefined contact geometries.
In this work, we introduce NeuralTouch, a multimodal framework that integrates NDFs and tactile sensing to enable accurate generalizable grasping through gentle physical interaction. Our approach leverages NDFs to implicitly represent the target contact geometry, from which a deep reinforcement learning policy is trained to refine the grasp using tactile feedback. This policy is conditioned on the neural descriptors and does not require explicit specification of contact types.
We validate NeuralTouch through ablation studies in simulation and with zero-shot transfer to real-world manipulation tasks, such as peg-out/in-hole and bottle lid opening. Our results show that NeuralTouch significantly improves grasping accuracy and robustness over baseline methods, offering a general framework for precise, contact-rich robotic grasp-based manipulation. Project webpage: \url{https://yijionglin.github.io/neuraltouch/}.

\end{abstract}


\section{INTRODUCTION}\label{sec:Intro}


A commonplace behaviour in humans is our ability to glance at an object to determine its general position and then use touch alone to grasp it with precision. For example, after seeing where a plug is located in a socket, we can perform the precise manipulations to unplug and replug it using just our sense of touch. This behaviour remains challenging to implement artificially in robotics,  and typically involves two phases: 1) an initial coarse phase where vision captures global information essential for contact-rich downstream tasks, and 2) and a subsequent fine phase where touch determines the optimal grasping pose, utilizing the prior visual information about the object's pose and geometry. 

However, current applications of vision and touch for robotic manipulation tend to be constrained by several factors. First, in common scenarios objects are already optimally positioned in the hand for grasping by the robot \cite{calandra2018regrasp}. Second, policies can be restricted to manipulating objects or contact features that were trained already, and so lack the ability to generalize to novel objects \cite{bauza2024simple}. Third, the independent use of vision and touch modalities can reduce their synergistic potential \cite{martin2023visualtactile}. Fourth, multimodal policies developed in simulation struggle to transition seamlessly to real-world environments. Here, we seek to address these challenges by proposing a novel multimodal policy learning framework capable of overcoming these limitations.

In this paper, we present \textit{NeuralTouch}, a tactile reinforcement learning (RL) framework to learn policies with neural descriptor fields (NDFs) \cite{simeonov2022neural}. Our goal is to improve the grasping accuracy of NDF-based methods with use of touch while maintaining generalizability to inter-category objects. Furthermore, our framework will not restrict the NDF-based tactile control to limited, predefined contact geometries. 

Experimentally, we focus on precise grasping with a tactile gripper through the aforementioned visual (coarse) and tactile (fine) phases. Specifically, in the coarse phase, we use an NDFs to generate a pre-grasping pose, then the fine phase focuses on in-hand tactile servo control of the gripper fingers that repositions and reorients the gripper to achieve a specific grasp. This process is challenging due to the need to interpret the underlying object geometry in combination with precise control of a 7-DoF robot arm and parallel jaw gripper. 

The main contributions of this work are as follows:\\
\noindent1) We propose a deep-RL-based framework with neural descriptor fields to train a general tactile policy that does not need any explicit assumption about prior contact geometry.\\
\noindent2) We demonstrate that our NeuralTouch strongly complements state-of-the-art vision-based grasping to achieve the desired grasping pose with improved accuracy.\\
\noindent3) We validate this experimentally with zero-shot sim-to-real policy transfer and few-shot demonstration to showcase that NeuralTouch solves a variety of downstream manipulation tasks over a variety of objects.

\section{RELATED WORK} \label{sec:related}

\subsection{Tactile Robot Control} \label{subsec:bi_drl_related}
The rapid development of soft, high-resolution visuotactile sensors \cite{lepora2021soft, abad2020visuotactile} has spurred various designs of tactile manipulators \cite{lu2025dexitac, jiang2025rotipbot}, making the design of effective controllers for tactile robots a prominent topic in robotics. The recent success of data-driven methods, such as deep learning and deep reinforcement learning (RL), has further motivated their application to challenging or novel domains, including high-DoF robot control~\cite{liu2025snake, yang2024anyrotate} and tactile robot control~\cite{church2020deep, amadio2019exploiting}. For example, \cite{jiang2023transparent} developed a learning framework using GelSight~\cite{yuan2017gelsight} to detect the state of transparent objects for precise grasping. More recently, \cite{han2025deformable} proposed a control framework based on transformers~\cite{vaswani2017attention}, leveraging high-resolution tactile sensing for safe and robust grasping of deformable objects.

Analogous to visual servo control, there are two main classes of tactile servo control: pose-based tactile control and image-based tactile control \cite{lepora2021pose}. For pose-based tactile servoing, the key idea is to explicitly estimate the contact pose between the tactile sensor and the object through tactile feedback, then use a simple PID controller to move towards a desired relative pose, which has been successfully applied to surface- and edge-following control tasks~\cite{lepora2019pixel, lepora2021pose, lepora2022digitac}. With an additional heuristic controller, these methods extend to object-pushing to a goal using tactile feedback only \cite{lloyd2021goal}. These tactile servoing methods extend to predict the contact-induced shear pose of the sensor to achieve shear-dependent robotic tasks, such as tracking a moving object~\cite{lloyd2024shear}. Similarly, \cite{wilson2023cable, she2021cable} achieve cable manipulation tasks by estimating the contact pose between a tactile sensor and a cable with Principal Component Analysis (PCA), using PID controllers with the contact pose as input. Some works have also explored the use of deep RL for learning tactile policies to control contact pose observations, for tubular objects \cite{zhao2023skill} and deformable linear object manipulation \cite{pecyna2022visual}.


While pose-based tactile control requires a pre-trained model for predicting the contact pose, image-based tactile control can be more straightforward when the controller outputs robot action directly with tactile images as input. This property aligns well with deep reinforcement learning to achieve end-to-end learning. Several works \cite{church_tactile_2021, lin2022tactilegym2, lin2023bitouch} apply deep RL to learn policies with tactile images for a suite of robotic tasks in simulation and successfully deploy them to real-world scenarios with zero-shot sim-to-real transfer. 
However, due to their reliance on low-dimensional action spaces, these approaches are constrained to the narrow scope of interacting with or tracking simplistic flat edges and surfaces. Such limitations inherently restrict the generalizability of these tactile systems to tasks requiring comprehensive 3D manipulation. Instead, the NeuralTouch method proposed here learns policies to fully control the robot with 6D poses and gripper finger distance, enabling a wider range of manipulation tasks.

\subsection{Neural Descriptors Fields for Robot Manipulation} 
Recent advances in neural fields have been highly successful in computer vision \cite{xie2022neural}, and some techniques, such as signed distance functions and 3D Gaussian Splatting, have been combined with tactile robotics for 3D shape reconstructions~\cite{comi2024touchsdf, comi2024snap}. Similarly, neural feature fields (or neural descriptor fields (NDFs)), as a subset of neural fields, can implicitly represent object geometric features at a part level, and the generalization to different objects across a category has incentivised robotics researchers to leverage it for object-centric robot manipulation. NDFs can represent the spatial relationship between a point/pose and an object with implicit descriptors~\cite{simeonov2022neural}, which are the concatenation of the activation value from each layer of a pre-trained occupancy network. Further, NDFs can facilitate robot manipulation with a few demonstrations specifying the desired relative pose between the end-of-effector and an object \cite{simeonov2023se, simeonov2023shelving}. Instead of requiring the entire object shape, local NDFs have been proposed that only focus on the desired local feature of a shape~\cite{chun2023local}, enabling more robust manipulation when the object is partly occluded. More recently, dynamic 3D descriptor fields that fuse the output from visual foundation models to represent objects' features have been proposed~\cite{wang2023d}, which enables zero-shot robot manipulation. 
Although these methods can attain a level of performance that generalizes well across different objects, their efficacy may be compromised in scenarios that demand precise contact pose. This limitation can arise from the absence of tactile feedback, which is crucial for the fine-tuning of the robot's movements for accurate task execution. In this paper, we will investigate how to combine tactile sensing and NDFs for generalizable and precise manipulation.

\subsection{Tactile Policies for Precise Manipulation Tasks} 
Tactile sensing plays a crucial role in precise robotic manipulation tasks. For instance, to balance an object during in-hand grasping it is essential for the robot to estimate the object's centre of mass through touch~\cite{wang2025regrasp}. A typical precise manipulation task that requires tactile feedback is object insertion, which is essential for applications such as assembly, manufacturing, and medical procedures. Most work on object insertion has leveraged shear deformation of the tactile skin for insertion adjustment because the shear induced by the contact between the object and the hole is the key tactile feature that detects the external contact state. Dong et al. \cite{dong2019tactile} used an LSTM model to predict the pose offset between parts and target position from tactile flow, but simplified the problem by discretizing the action space, which applies only to simple insertion tasks. In their subsequent work \cite{dong2021insertion}, they treated this task as a continuous control problem and learned a deep RL policy with a tactile gripper that can insert several types of objects of simple geometries in an end-to-end manner. Similarly, Zhao et al. \cite{zhao2023skill} presented a tactile sim-to-real RL framework for tube insertion by estimating the object pose. Freud et al.~\cite{freud2025simshear} proposed using a GAN to generate synthetic real tactile images with shear effects from simulated tactile images, which can then be used for simple surface-following tasks with PID controllers. Extrinsic contacts have also been studied to solve the insertion task \cite{higuera2023perceiving}.

In work more related to our present study, Bauza et al.~\cite{bauza2024simple} developed a bimanual robotic system with vision and touch for object pick-and-place. Their system can re-grasp an object to improve pose accuracy for insertion by predicting the in-hand object pose with a supervised-learning model \cite{bauza2023tac2pose, villalonga2021tactile}. The precise estimation of the object pose enables it to solve this task directly without further interaction during insertion. However, this approach relies on an expensive dual-arm setup, 
making it less practical for real-world applications compared to our method, which utilizes a more efficient and compact single-arm system without compromising performance. Also, they only considered 2D initial configurations of the objects which lie on a table, while we consider objects that can be placed with random 6D poses thanks to the generalizability of NDFs. Note that their method used supervised learning with known object shapes for pose estimation, while our method does not need the explicit object models thanks to the generalizability of NDFs.



\begin{figure*}
\vspace{-2em}
  \centering
     \includegraphics[width=0.99\linewidth]{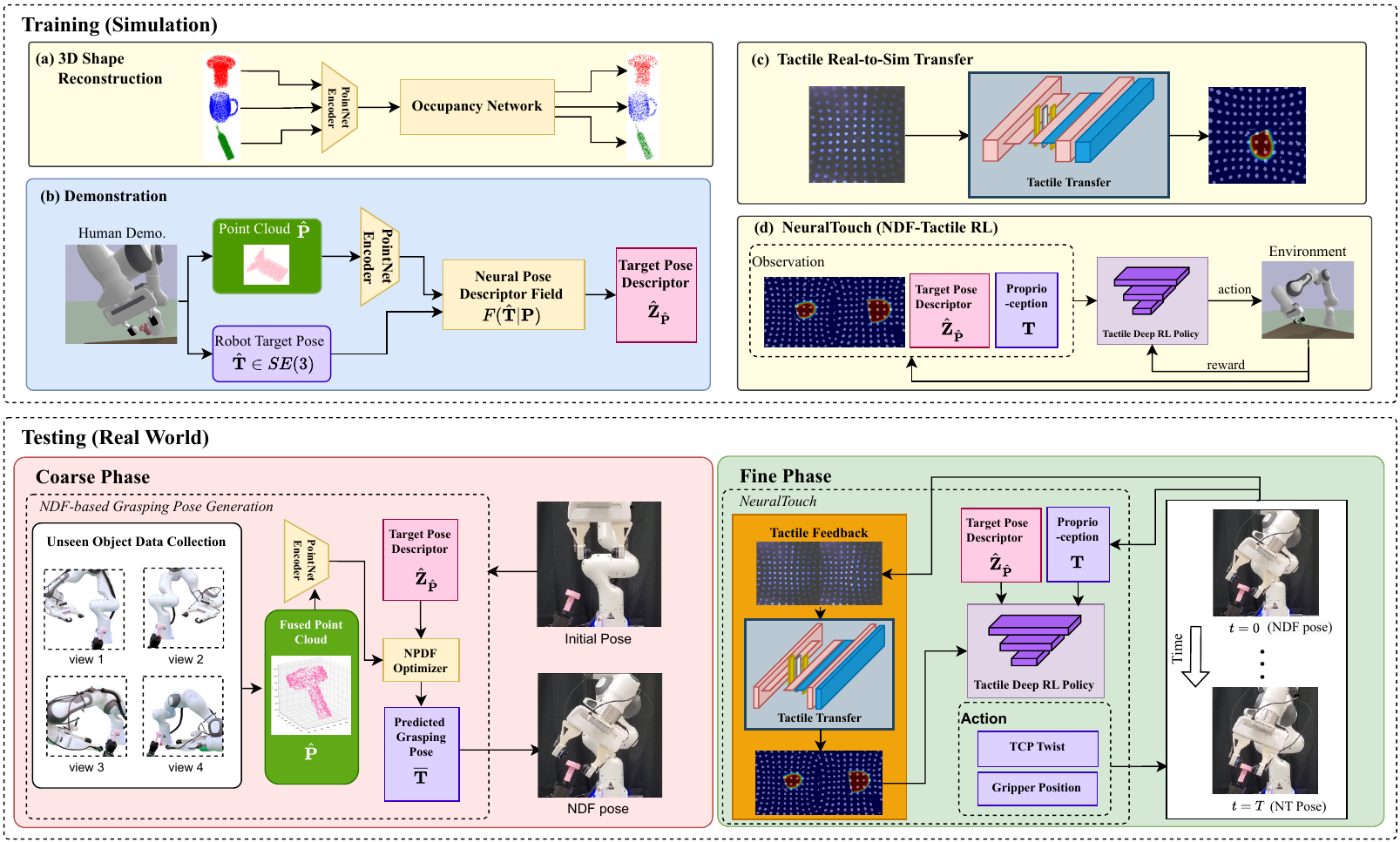}

    \caption{Overview of the NeuralTouch: In simulation, we first pre-train an occupancy network which is the core component of the Neural Pose Descriptor Fields. Secondly, we collect human demonstrations along with object point clouds and robot target grasping pose descriptors depending on the manipulation tasks. Thirdly, we train an RL policy with tactile and proprioceptive feedback, to achieve fine grasping poses implicitly specified by these collected descriptors. After obtaining the NPDF and a well-trained policy, our system is directly deployed in the real world with a real-to-sim tactile transfer to accurately grasp unseen objects, executing manipulation tasks such as unplugging a bolt-like USB and inserting it into a socket.}
  \label{fig:framework}
\end{figure*}
\section{Methods} \label{sec:method}
We separate the robotic grasping task into two phases: a coarse vision-guided phase and a fine tactile-guided phase. Note that while we structure this task similarly to other coarse-to-fine approaches~\cite{johns2021coarse}, we do not rely on any specific methods from those approaches. In the coarse phase, we leverage the descriptor generated from NDFs to calculate the coarse target grasping pose. Then, in the fine phase, we apply a tactile RL policy to accurately grasp an object with a desired contact pose represented by the NDFs descriptor.

Specifically, we focus on learning a tactile RL policy that can be generalized to different target contact poses for different objects or tasks with the help of implicit neural descriptors from NDFs. The tactile RL policy should not only consider the local contact to achieve safe, gentle contact but also have a sense of its desired contact pose with respect to the global shape of an object. Our method consists of three modules:\\
\noindent1) A PointNet Encoder \cite{qi2017pointnet} with Neural Pose Descriptor Fields \cite{simeonov2022neural} that learns implicit descriptors for various object shapes. These implicit representations describe the geometric relationships between poses (local frames) and the corresponding local shapes of inter-category objects.\\ 
\noindent2) A module to generate an initial coarse grasping pose using regression over the NDFs \cite{simeonov2022neural}. \\
\noindent3) A NeuralTouch RL module that learns a general tactile robotic policy conditioned on the implicit neural descriptors to achieve the desired fine grasping pose while maintaining safe, gentle physical interaction between the tactile robot and a manipulated object, given tactile and proprioceptive feedback.

\subsection{Background: Neural Descriptor Fields}\label{subsec:hardware}
The key idea of the Neural Descriptor Field is to implicitly represent geometric correspondences \cite{simeonov2022neural}. Specifically, an NDF descriptor represents the spatial relationship between a point $p$ and an object $o$ using a function $f$ parameterized as a neural network that maps the point coordinate $\mathbf{x} \in \mathbb{R}^3$ and the object point cloud $\mathbf{P\in\mathbb{R}^{3 \times N}}$ to a descriptor $z\in \mathbb{R}^{d}$,
\begin{equation} 
f(\mathbf{x} \mid \mathbf{P}): \mathbb{R}^3 \times \mathbb{R}^{3 \times N} \rightarrow \mathbb{R}^{d}.
\label{eq:NDFs_RL}
\end{equation}
As such, the NDFs has two appealing properties:

\noindent{\em 1) SE(3)-Equivariance:}
As an NDF is constructed with a rotational SO(3)-equivariant neural network structure \cite{deng2021vector}, and its input is aligned to the point cloud centre, it is SE(3)-equivariant with the mapping satisfying
\begin{equation} 
f(\mathbf{x} \mid \mathbf{P}) \equiv f(\mathbf{T} \mathbf{x} \mid \mathbf{T} \mathbf{P}).
\label{eq:NDFs}
\end{equation}
This property ensures the spatial descriptors are invariant to different poses of the same object.

\noindent{\em 2) Geometric Correspondence:}
Another appealing property of NDFs is that they learns a geometric correspondence over inter-category objects, as their backbone is trained for category-level shape reconstruction \cite{mescheder2019occupancy}. For example, the descriptors of points near the handles of two different mugs will be similar because the handles share a similar ring shape. 

Based on property (1), NDFs can represent 6D poses $\mathbf{T}\in SE(3)$ instead of just point positions, by stacking the descriptors of the individual points in a set of non-collinear query points $ \left \{ \mathbf{x}_i \in \mathcal{X}  |\mathcal{X} \in \mathbb{R}^{3 \times N_{q}},  i = 1, 2, \cdots , N_q  \right \}$ with rigid configuration,  where the number of the points $N_q \geq 3$. Thus, an NDF can represent a pose $\mathbf{T}$ by rigidly transforming the query points $\mathcal{X} $ to the coordinates of $\mathbf{T}\mathcal{X}$:
\begin{equation} 
\mathcal{Z}=F(\mathbf{T} \mid \mathbf{P})=\bigoplus_{\mathbf{x}_i \in \mathcal{X}} f\left(\mathbf{T x}_i \mid \mathbf{P}\right),
\label{eq:NDFs}
\end{equation}
\noindent where $F$ maps a point cloud $\mathbf{P}$ and a pose $\mathbf{T}$ to a \textit{pose descriptor} $\mathcal{Z} \in \mathbb{R}^{d \times N_q}$. Note that $F$ inherits both properties (1) and (2) above.

It has been shown~ 
\cite{simeonov2022neural} that a desired grasping pose $\mathbf{T}_g$ for an unseen object (with points cloud $\mathbf{P}_u$) can be calculated by minimizing the distance between the descriptor of a pose $\mathbf{T}$ and the descriptor of the target pose $\mathbf{T}_d$ for an object (with points cloud $\mathbf{P}_d$), such that\footnote{In practice, a target pose descriptor is the average of multiple pose descriptors given in several demonstrations.}
\begin{equation} 
\mathbf{T}_g=\underset{\mathbf{T}}{\operatorname{argmin}}\|F(\mathbf{T} \mid \mathbf{P}_{u})-F(\mathbf{T}_d \mid \mathbf{P}_d)\|.
\label{eq:NDFs}
\end{equation}
This quantity is used for the initial coarse grasping pose generation process with pose regression using NDFs.

\subsection{Tactile RL Policy with the Neural Pose Descriptor}\label{subsec:hardware}
While the NDF is sufficient for generating coarse grasping poses, the actual derived pose can be inaccurate due to the differences between objects, imperfect camera calibration, and incomplete point clouds. Such inaccuracy hinders the direct application to tasks where precision is key to achieving success, such as peg-in-hole. Also, as the method lacks tactile sensing, it cannot achieve gentle grasping, which can potentially damage the robot or object. Therefore, we aim to solve this challenge with \textit{tactile servoing} to accurately and safely achieve target contacts for grasping \cite{lepora2021pose, church_tactile_2021}. 

However, previous works on tactile servo control \cite{lepora2020optimal, lepora2021pose, church_tactile_2021,lin2022tactilegym2} cannot be applied directly because they either design or learn a tactile policy $\pi^{\text{G}_\tau}$ for a single target contact pose $\mathbf{T}^{\text{G}_\tau}$, where $\tau$ is a type of predefined contact geometry (e.g. a flat edge) and $\text{G}_\tau$ is a predefined target contact configuration for that type of geometry (e.g. normal contact to a flat edge). Thus, the robot action is
\begin{equation} 
a = \pi^{\text{G}_\tau}(i^\text{c}, e),
\label{eq:NDFs_RL}
\end{equation}
where $i^\text{c}$ denotes a tactile image of the contact and $e$ denotes the proprioceptive feedback of the robot (usually the pose of the robot end-effector).

To relax this assumption, we propose a novel method, \textit{NeuralTouch}, for learning a general tactile servoing policy that is conditioned on a neural pose descriptor $\mathcal{Z}^{\text{G}_\tau}$:
\begin{equation} 
a = \pi(i^\text{c}, e, \mathcal{Z}^{\text{G}_\tau}).
\label{eq:NDFs_RL}
\end{equation}
Here $\mathcal{Z}^{\text{G}_\tau}$ represents the geometric correspondence, to provide sufficient information for a tactile policy to understand the target contact geometry between an object and the tactile sensors, along with tactile images and proprioceptive feedback.

\subsection{Reward Design}
The objective of the RL training is to reach the target grasping poses (see Sec.~III-C for details) based on tactile feedback, with or without the Neural Descriptor. Specifically, the reward function $R_{t}$ can be expressed at time step $t$ as:
\begin{equation} 
R_{t} = w_1\sum_{i=1}^{2}\left \| p_{t}^{\rm g_i} - p_{t}^{\rm f_i} \right \| + w_2S(q^{\rm g}_{t},q_{t}^{\rm e})+r_{\rm stable}+r_{\rm act}+r_{\rm term},
\label{eq:reward}
\end{equation}
where $w_{j}<0$ $(j\in\{1,2\})$ are reward weights; $p_{t}^{\rm g_1}$ and $p_{t}^{\rm g_2}$ denote the target positions of the left and right fingers for the current goal grasping pose, respectively; $q^{\rm g}_{t}$ represents the orientation of the current target grasping pose; $p_{t}^{\rm f_1}$ and $p_{t}^{\rm f_1}$ are the positions of the left and right gripper fingers, respectively; and $q^{\rm e}_{t}$ denotes the current orientation of the robot end-effector. The function
$
S(\boldsymbol{\Phi}, \boldsymbol{\Psi})
= \sum_{k=1}^{3} \left( 1 - \cos(\phi_k - \psi_k) \right)
$
measures the sum of cosine distances between the corresponding orientation angles
$\phi_k$ and $\psi_k$, where $\boldsymbol{\Phi} = [\phi_1,\phi_2,\phi_3]^\top$
and $\boldsymbol{\Psi} = [\psi_1,\psi_2,\psi_3]^\top$ denote the Euler-angle
representations of the two orientations.

In addition, we include a stability reward $r_{\rm stable}>0$ that is activated when the robot reaches the desired grasp pose and maintains it for more than 50 time steps, encouraging sustained and stable grasping. We also have an action penalty $r_{\rm act}<0$ term to improve the learned policy stability, smoothness, and energy efficiency. Finally, an early termination penalty term $r_{\rm term}<0$ is applied if tactile contact is not detected for more than 100 time steps, discouraging non-contact behaviours and improving training efficiency. 

\subsection{Unified Semantic Grasping Poses}\label{subsec:hardware}
Given our coarse to fine grasping process is based on tactile servo control, our NeuralTouch framework should be able to find a specific target contact $G^{\tau}$ specified by a corresponding neural pose descriptor $\mathcal{Z}^{\tau}$. However, this is impractical to implement during training as the state space of NDFs is too large for efficient exploration. Instead, we can make a simplifying assumption that the target contact between the gripper and object is where the tactile sensors are perpendicular to the local surface of the object, which holds in situations where we require a robust grasp that maintains normal contact with the local surface. We can then unify different $\mathcal{Z}$ descriptors that are close to $\mathcal{Z}^{\tau}$ with the same target contact $G^{\tau}$ conditioned on the local geometry. For example, if the target is to grasp a mug on its rim, the target contact maintains the contact normal between the tactile sensors and the local surface of the rim with the gripper pose parallel to the mug pose.

Different contact features can be determined by different dimensionalities of the contact parameters. For example, a flat 3D surface relies on three parameters: contact depth, roll, and pitch, whereas a flat 3D edge requires two additional parameters: contact offset and yaw (see \cite{lepora2020optimal} for a more detailed explanation). In this work, we consider not only these two contact geometries but also curved surfaces and handles that require 6 parameters of a full 6D contact pose. 

\section{Experimental Design}\label{sec:exps}

Our experiments are designed to demonstrate the effectiveness and generalizability of the proposed NeuralTouch framework. In particular:
1) Are the NDFs descriptors informative enough for one single RL policy to learn to grasp different contact features of different objects?
2) Does NeuralTouch improve the grasping accuracy on various unseen objects compared to the other SOTA baseline methods on contact-rich manipulation tasks requiring high accuracy?
3) What is the NeuralTouch zero-shot sim-to-real performance?

\subsection{Robotic System with Vision and Touch}
\subsubsection{Hardware}
Our robotic system comprises a 7-DoF Franka Panda arm with a wrist-mounted Intel RealSense Depth Camera D435 (shown in Fig.~\ref{fig:framework}, bottom). Unlike the original NDF study that used four cameras for object point cloud collection~\cite{simeonov2022neural}, we simplify the system to use just a single depth camera. This eye-in-hand depth camera is extrinsically calibrated and captures point clouds from four distinct views relative to the robot’s base frame that are subsequently fused together. Each finger is equipped with a customized compact version of the TacTip sensor for the Franka gripper. 
\subsubsection{Simulation}
In simulation, we integrate Tactile Gym 2.0 for tactile reinforcement learning~\cite{lin2022tactilegym2} with the NDF simulated environment, using PyBullet\cite{coumans2016pybullet}. The simulated robot represents a 7-DoF Franka Panda arm equipped with the 3D models of our two tactile sensors. The camera setup in the simulation is the same as detailed in reference \cite{simeonov2022neural}.
\subsubsection{Real-time Control Architecture and Computational Performance}
\begin{figure}[t]
  \centering
  \includegraphics[width=1\linewidth]{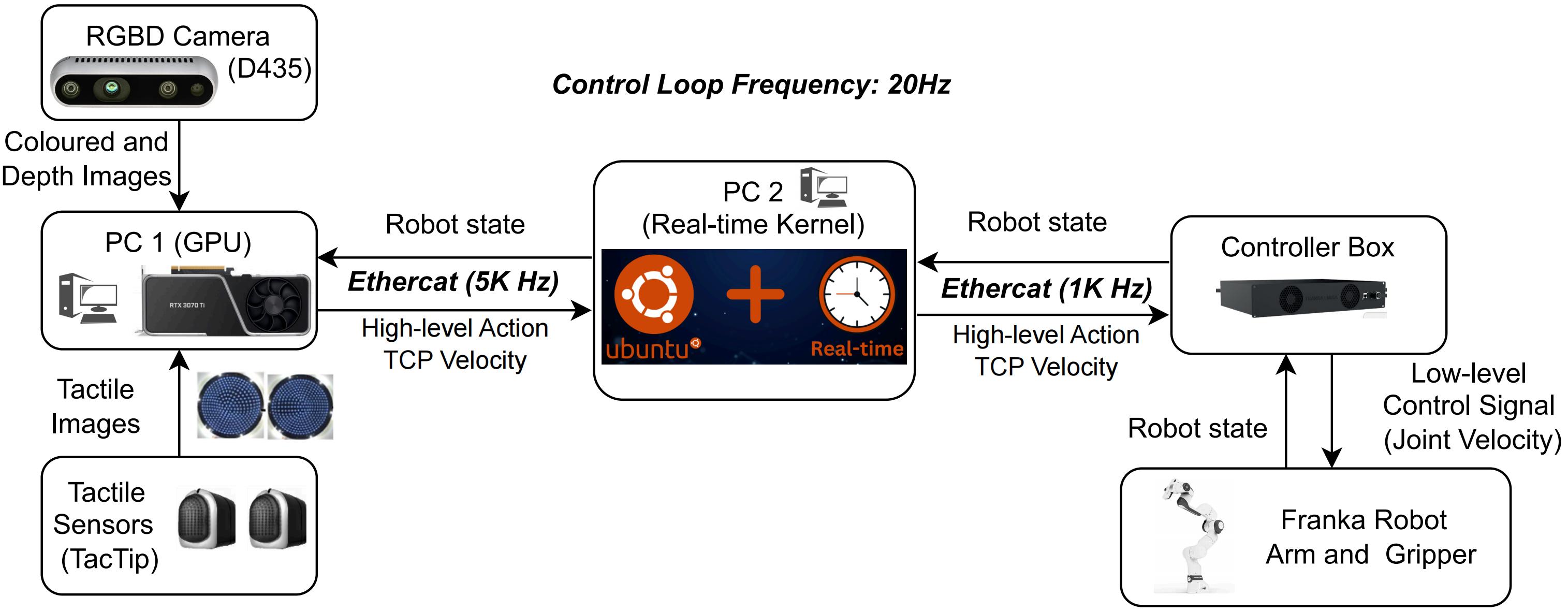}
  \vspace{-0.5em}
    \caption{Overview of the real-time tactile robotic control system.
    }
  \label{fig:neuraltouch_control_framework}
\end{figure}

The proposed robotic system comprises three neural model components: (i) NDF-based target pose inference, (ii) real-to-sim tactile translation using GANs, and (iii) RL policy execution. As shown in Fig.~\ref{fig:neuraltouch_control_framework}, the system is implemented across two PCs. The NDF optimisation is executed once per episode (or task reset) to compute a target grasp or interaction pose, and is not part of the real-time control loop; it typically converges within 1 s on a single GPU. During task execution, two real-to-sim GANs and the RL policy run in parallel, with the overall control cycle bounded by the maximum per-step inference latency ($\approx 45~\mathrm{ms}$), enabling stable closed-loop control at 20 Hz. This setup supports GPU-accelerated perception and policy inference while maintaining reliable real-time robot control.

\begin{figure}[t]
  \centering
  \includegraphics[width=0.8\linewidth]{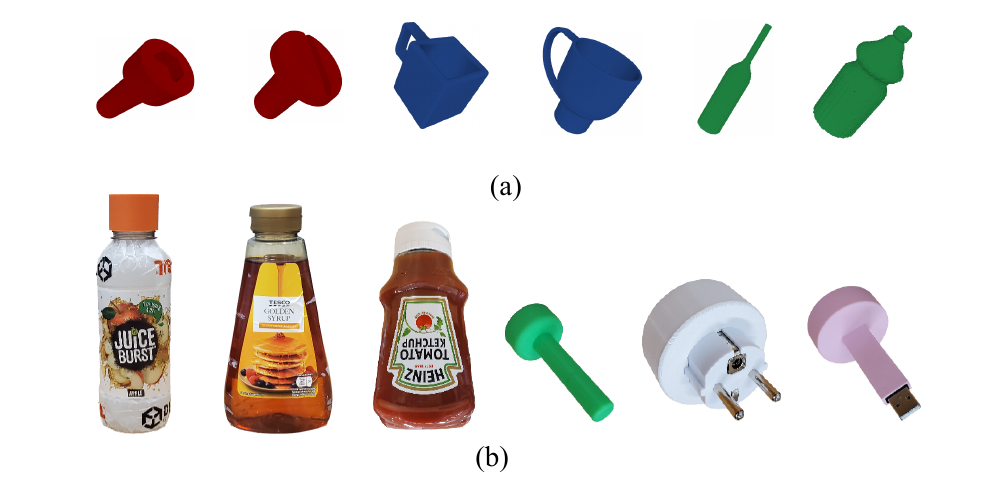}
  \vspace{-0.5em}
    \caption{Simulation and real-world objects used in this paper. (a) Representative simulated objects used for occupancy network training, including bolts, mugs, and bottles. (b) Real-world objects used for evaluating the proposed manipulation tasks, from left to right: an apple juice bottle, a syrup container, a ketchup bottle, a bolt, a plug adaptor, and a USB connector.} 
  \label{fig:sim_obj}
\end{figure}

\subsection{Tasks Setup}\label{subsec:task}
First, we design an ablation study and compare our method to four baselines: NDFs \cite{simeonov2022neural}, NDFs+RL-Touch, Coarse--to-Fine Imitation Learning (C2FIL)~\cite{johns2021coarse}, and C2FIL+RL-Touch (this latter baseline trained under RL-Touch is detailed in Sec.~\ref{subsubsec:rl_policy}). Specifically, we analyze grasping accuracy by measuring position errors and orientation errors for various target features of different objects in simulation.

To further evaluate our proposed method, we consider two tasks both in simulation and in reality, with three phases: 1) a coarse phase where the robot uses vision to locate and approach the unknown target feature pose of an object; 2) a fine phase where the robot uses in-hand tactile servoing of the gripper fingers to achieve a precise grasping pose; and 3)~a replay phase where the robot executes a predefined skill to complete the task. The fine phase is particularly challenging due to the need for 7-DoF robot control and an understanding of the object's geometry to reposition and reorient the robot.

\subsubsection{Simulation-based Object Pick-and-Place}
The robot aims to grasp a mug by its rim or horizontal handle, or a bottle by its neck, and place it horizontally on a table. The task is considered successful if the robot successfully picks up the object and places it on the table in a stable upright position. The object is initialized with a random pose above the table. The first contact is detected in simulation during the placing process, which releases it from the gripper (using the default Pybullet's method for contact detection).

\subsubsection{Simulation-based Bolt out/in Hole}
The robot aims to locate and unplug a bolt from one hole and insert it into another. The poses of the bolt and the first hole are both unknown and initialized randomly above the table, while the pose of the second hole mounted on the table is known. 

\subsubsection{Real Bottle Lid Opening}
The robot aims to locate a bottle placed above the table and open its lid. The opening action of rotation is given by demonstration, hence the robot needs to grasp the lid with the correct pose to successfully execute the downstream task. 

\subsubsection{Real Bolt out/in Hole}
This task is similar to that in simulation but with three out-of-distribution objects of increasing difficulty due to higher peg-hole clearance: a bolt, an adaptor plug, and a USB with a bolt-shaped adaptor. The yaw angle for the plug and USB is fixed, and the other components of their initial poses are sampled randomly. The clearance of the hole for inserting the green bolt is 2\,mm, the plug is 1\,mm, and the USB is 0.5\,mm. The widths of the bolt, USB, and plug are 34\,mm, 48\,mm, and 60\,mm respectively.

The most challenging task, bolt out/in hole, has been well studied in relation to the physical interaction of inserting a peg already grasped by a tactile robot \cite{zhao2024fots, xu2022efficient, higuera2023extrinsic, kim2022active, dong2021insertion}. In contrast, our focus is on grasping the peg with a desired grasping pose using in-hand tactile serving. Our method enables smooth insertion into a hole without requiring further physical adjustments during the insertion process. 

\begin{figure}[t]
  \centering
    \includegraphics[width=1\linewidth]{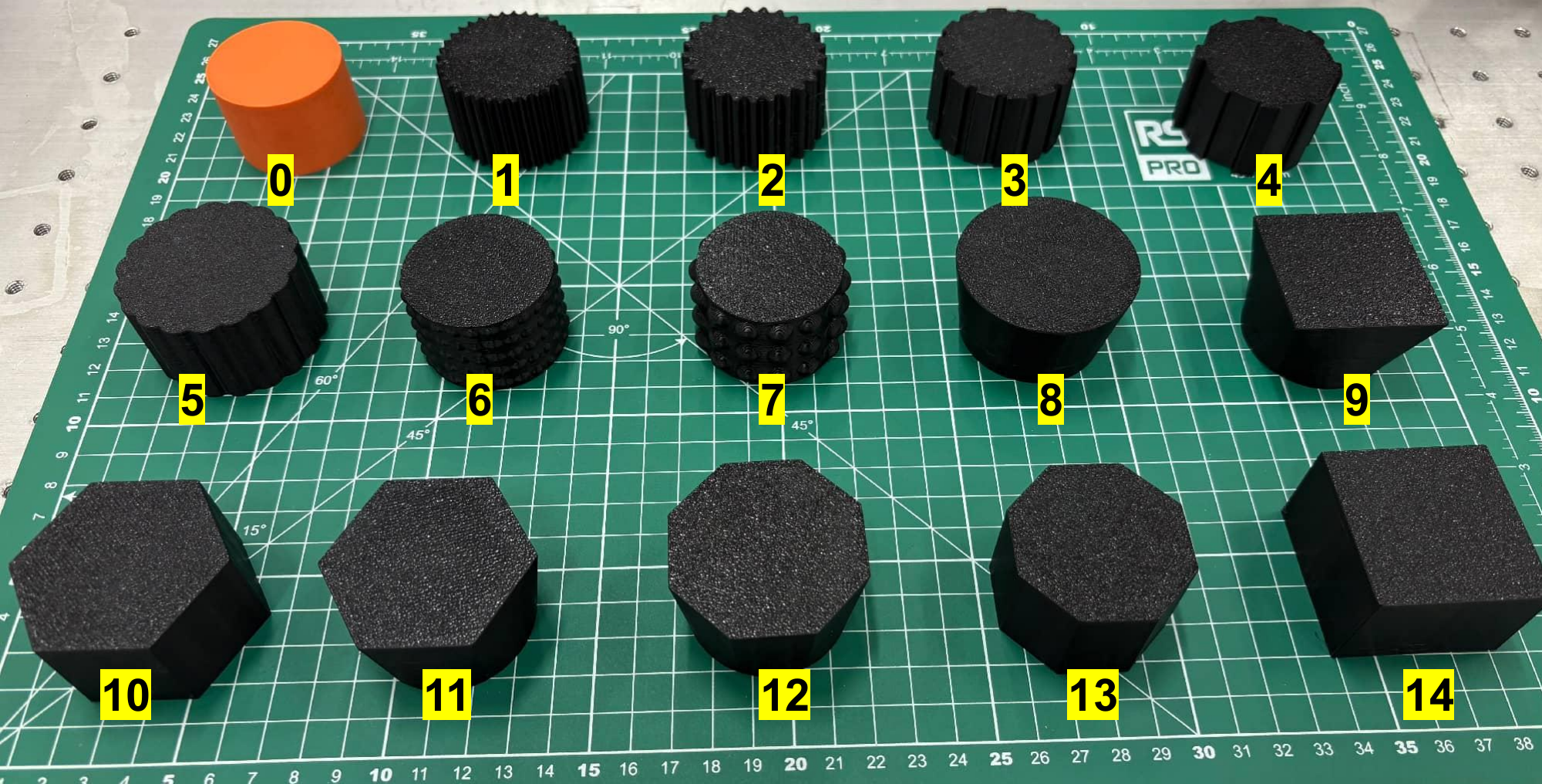}
    \vspace{-.5em}
    \caption{A diverse set of bottle lids with varying geometries and surface textures used to evaluate the generalisation capability of the proposed method. The lids include cylindrical, polygonal, and irregular shapes with smooth, ridged, and patterned side surfaces, enabling systematic assessment of robustness to shape and texture variations in the bottle-lid-opening task.}
  \label{fig:bottle_lids}
\end{figure}

In addition, to further evaluate the generalisation ability of our method, we conducted the bottle-lid-opening tasks with lids of various shapes and textures, as shown in Fig. \ref{fig:bottle_lids}. The results are reported in Sec.~\ref{subsec:real_world_exp_results}.


\begin{table*}
\centering
\addtolength{\tabcolsep}{-5pt}
\caption{AVERAGE PERFORMANCE OF DIFFERENT METHODS TESTED WITH 6 TARGET FEATURES OF DIFFERENT OBJECTS. ‘T’ denotes RL-Touch.}
\label{table:average_error}
\begin{tabular}{>{\hspace{0pt}}m{0.077\linewidth}|>{\hspace{0pt}}m{0.073\linewidth}|>{\hspace{0pt}}m{0.069\linewidth}|>{\hspace{0pt}}m{0.073\linewidth}|>{\hspace{0pt}}m{0.069\linewidth}|>{\hspace{0pt}}m{0.073\linewidth}|>{\hspace{0pt}}m{0.069\linewidth}|>{\hspace{0pt}}m{0.073\linewidth}|>{\hspace{0pt}}m{0.069\linewidth}|>{\hspace{0pt}}m{0.073\linewidth}|>{\hspace{0pt}}m{0.069\linewidth}|>{\hspace{0pt}}m{0.073\linewidth}|>{\hspace{0pt}}m{0.069\linewidth}} 
\hline
target feature & \multicolumn{2}{>{\hspace{0pt}}m{0.142\linewidth}|}{mug rim} & \multicolumn{2}{>{\hspace{0pt}}m{0.142\linewidth}|}{mug wall} & \multicolumn{2}{>{\hspace{0pt}}m{0.142\linewidth}|}{right-angle handle} & \multicolumn{2}{>{\hspace{0pt}}m{0.142\linewidth}|}{horizontal handle} & \multicolumn{2}{>{\hspace{0pt}}m{0.142\linewidth}|}{bottle head} & \multicolumn{2}{>{\hspace{0pt}}m{0.142\linewidth}}{bolt head} \\ 
\hline
average error & Position Error & Cosine Error & Position Error & Cosine Error & Position Error & Cosine Error & Position Error & Cosine Error & Position Error & Cosine Error & Position Error & Cosine Error \\ 
\hline
NT (Ours) & \textbf{\textbf{0.8mm}} & \textbf{\textbf{0.0006}} & \textbf{\textbf{0.7mm}} & \textbf{\textbf{0.0007}} & \textbf{\textbf{0.9mm}} & \textbf{\textbf{0.0008}} & \textbf{\textbf{1.0mm}} & \textbf{\textbf{0.0005}} & \textbf{\textbf{0.9mm}} & \textbf{\textbf{0.0006}} & \textbf{\textbf{0.7mm}} & \textbf{\textbf{0.0005}} \\ 
\hline
NDFs~\cite{simeonov2022neural} & 13.6mm & 0.0083 & 12.4mm & 0.0039 & 11.0mm & 0.0091 & 11.1mm & 0.0086 & 9.0mm & 0.0069 & 12.5mm & 0.0072 \\ 
\hline
NDFs+T~ & 15.3mm & 0.0030 & 17.3mm & 0.0043 & 18.7mm & 0.0060 & 13.9mm & 0.0050 & 2.0mm & 0.0048 & 2.6mm & 0.0027 \\ 
\hline
C2FIL~\cite{johns2021coarse} & 17.3mm & 0.0052 & 16.9mm & 0.0054 & 16.6mm & 0.0046 & 15.4mm & 0.0074 & 20.2mm & 0.0051 & 3.4mm & 0.0018 \\ 
\hline
C2FIL+T & 22.5mm & 0.0034 & 25.1mm & 0.0040 & 20.8mm & 0.0061 & 24.8mm & 0.0052 & 15.2mm & 0.0038 & 2.2mm & 0.0013
\end{tabular}
\label{table:average_error}
\end{table*}
\subsection{Data Collection and Models Training}
\subsubsection{Neural Descriptor Fields} First, we first train an occupancy network \cite{mescheder2019occupancy} in simulation to reconstruct three classes of objects: bottles, mugs, and bolts. Training details described in \cite{simeonov2022neural} are used, and ShapeNet meshes~\cite{chang2015shapenet} utilized for the bottle and mug classes. We include additional mug objects that contain horizontal and right-angle handles and have also developed a custom mesh dataset for the bolt class.
After obtaining a well-trained occupancy network, the Neural Pose Descriptor Fields are constructed by concatenating all activations output across the network layers~\cite{simeonov2022neural}, which allows us to represent an object as a function mapping a 6D pose relative to the object to a spatial descriptor. We collect 12 NDFs vectors for each target feature to train the NeuralTouch RL policy.

\subsubsection{RL Policies}\label{subsubsec:rl_policy}
Proximal Policy Optimization (PPO)~\cite{schulman2017proximal}, an on-policy model-free deep-RL algorithm,  is used to train an RL policy in simulation for all objects described above. Specifically, we use the Stable-Baselines-3 \cite{raffin2019stable} implementation of PPO. Our NeuralTouch RL policy takes target NDFs vectors, tactile images, and proprioceptive feedback (end-effector pose and gripper fingers distance) as input, with output action space a 7-dimensional representation of the end-effector twist; i.e. the translational velocity $v \in \mathbb{R}^3$ and angular velocity  $\omega \in \mathbb{R}^3$, and the gripper finger distance $g_d\in \mathbb{R}$. The low-level controller converts the twist output from this policy to joint velocity targets via the inverse kinematics. The robot gripper is assumed to already be around the object with a random initial pose that mimics an inaccurate grasping pose from the NDF optimizer, sampled from the pose range $\left[\pm 20\,\text{mm},\pm 20\,\text{mm},\pm 20\,\text{mm},\pm 20\degree,\pm 20\degree,\pm 20\degree\right]$. Initial grasping poses are not sampled from the NDF optimizer during training due to the prohibitive computational expense.

To verify the effectiveness of NeuralTouch, we also trained an alternative `RL-touch' policy for comparison in our ablation study, which uses only tactile images and proprioceptive feedback as inputs without any neural descriptors from NDFs. In other words, this baseline tactile RL policy operates without any visual information about the object.

\subsubsection{Tactile Real-to-Sim Transfer}
To enable zero-shot sim-to-real policy transfer for NeuralTouch, we utilize a pix2pix GAN \cite{isola2017image} for tactile image transfer (with training details as in \cite{church_tactile_2021, lin2022tactilegym2}). 

We collected cylinder-feature data with pose ranges: $(x,y,R_x,R_y,R_z)\in[\pm 10\,{\rm mm}, \pm 6\,{\rm mm},\pm20^{\circ},\pm 20^{\circ},\pm 45^{\circ}]$ and $z\in[0,4.5]$\,mm, with a training dataset for each tactile sensor. Each training dataset comprises 5000 tactile images and the validation dataset has 2000 tactile images collected both in simulation and with a physical Franka robot on paired random contacts. Samples are labelled with the relative poses between the sensor and cylinder. 

\subsection{Model Implementation Details and Parameter Choices}
To ensure reproducibility and a fair comparison with prior work, we adopt established hyperparameter settings for the baseline methods while explicitly specifying the design choices for our proposed approach. For the NDF module, we use the hyperparameters reported in the original paper~\cite{simeonov2022neural}; this included those for the occupancy network architecture, optimisation procedure, and inference settings to preserve the intended behaviour of the NDF when integrated into RL. For the real-to-sim tactile translation GAN and the RL policy learning used in the baseline methods, we follow the hyperparameter configurations reported in the original paper~\cite{lin2023bitouch} that have been validated on closely-related tactile manipulation tasks. For the coarse-to-fine imitation learning (C2FIL) baseline, we likewise use the parameter settings reported in the original C2FIL work~\cite{johns2021coarse} without additional task-specific tuning.

The NeuralTouch RL policy proposed in this work takes tactile images, robot proprioception, and the neural descriptor as inputs. The proprioceptive features and neural descriptor are concatenated and processed by a three-layer MLP with hidden dimensions [512, 256, 128], while the tactile images are encoded using a CNN with the same architecture as that used in \cite{lin2023bitouch}. The resulting feature embeddings are then concatenated and passed through a two-layer MLP with hidden dimensions [256, 128] to produce the final policy output. All remaining training and optimisation hyperparameters for NeuralTouch are kept identical to those used in \cite{lin2023bitouch}, ensuring that performance differences arise from the proposed representation and policy structure rather than from task-specific tuning.


\section{Experimental Results}\label{sec:exps_results}

\subsection{Ablation Study}\label{subsec:ablation}
\begin{figure}[b!]
  \centering
  \includegraphics[width=0.49\linewidth,trim={0 20 0 360},clip]{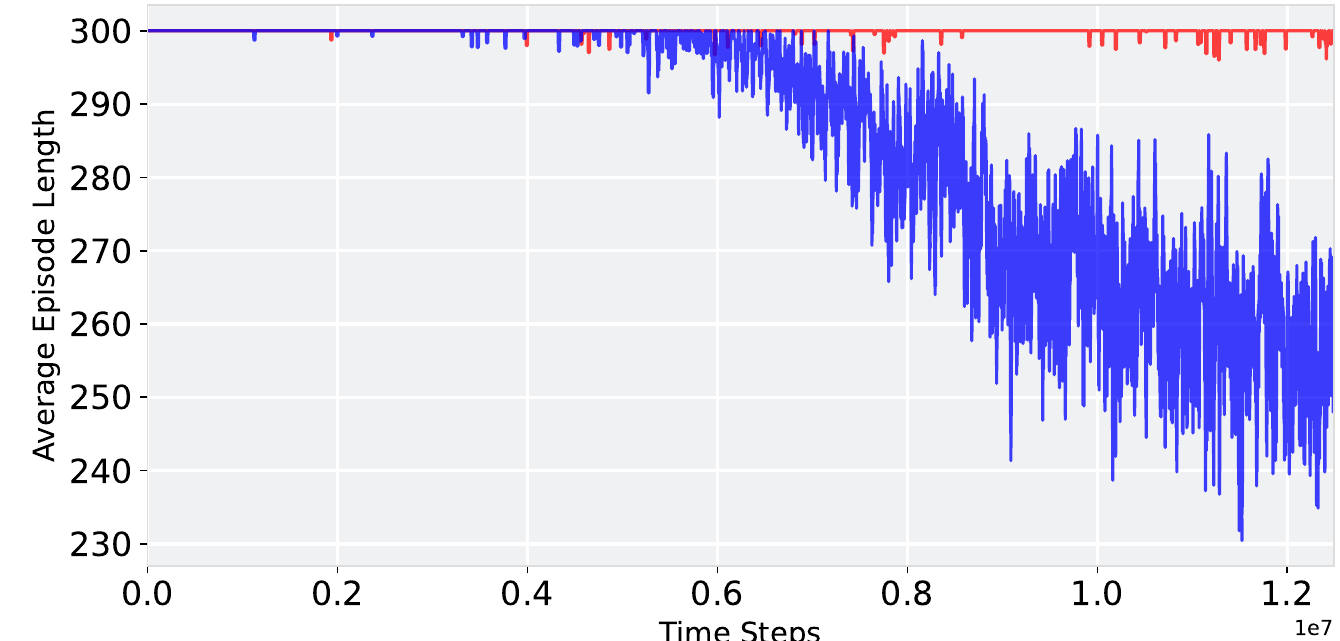}
  \includegraphics[width=0.49\linewidth,trim={0 350 0 0},clip]{pic/neuraltouch_rl_results.pdf}
    \caption{Comparison of training performances for NeuralTouch (blue) and vanilla RL-Touch (red) across 10 random seeds. (a) NeuralTouch, leveraging neural descriptors from NDFs, achieves superior asymptotic performance (higher average episode rewards), whereas vanilla RL-Touch struggles. (b) NeuralTouch completes tasks with significantly shorter episode lengths, demonstrating enhanced efficiency compared to RL policies without NDFs.} 
  \label{fig:learning_curve}
\end{figure}

We trained two tactile policies (RL-Touch and NeuralTouch) in simulation for 6 target features on three kinds of objects (Fig.~\ref{fig:sim_obj}): the mug rim, the mug wall, the mug right-angle handle, the mug horizontal handle, the bottle lid, and the bolt head. The RL-Touch policy has worse performance compared to the NeuralTouch policy (training curves shown in Fig.~\ref{fig:learning_curve}), which we attribute to the lack of visual information leading to ambiguity when relying on touch only. Thus, the NeuralTouch policy achieves optimal performance on different target features compared to the baselines, showing the effectiveness and efficiency of considering the implicit descriptor from NDFs. 

The experimental results of the ablation study conducted in simulation are reported in Table~\ref{table:average_error}. While the vanilla NDF is able to approximately reach the target poses, its performance is limited by the absence of fine-grained tactile feedback, resulting in relatively poor accuracy.

Interestingly, although NDF+RL-Touch performs better than the vanilla NDF when reaching the bottle-lid and bolt-head features, it achieves similar or even worse performance on the remaining four target features. This degradation arises because, without the visual guidance provided by NDF, the policy cannot reliably distinguish which feature is the intended target. In particular, the mug rim and mug wall, as well as the right-angle handle and horizontal handle, exhibit highly similar contacts during interaction, leading to feature ambiguity.

C2FIL and C2FIL+RL-Touch perform well on the bolt tasks due to the high geometric similarity among objects within the bolt category. However, their performance degrades substantially in object categories that contain instances with diverse shapes, highlighting the limited generalisability of C2FIL-based methods. In contrast, NeuralTouch jointly leverages visual and tactile modalities during physical interaction, achieving significantly higher grasping accuracy and stronger generalisation across object categories.

\begin{figure}
  \centering
  \includegraphics[width=1\linewidth]{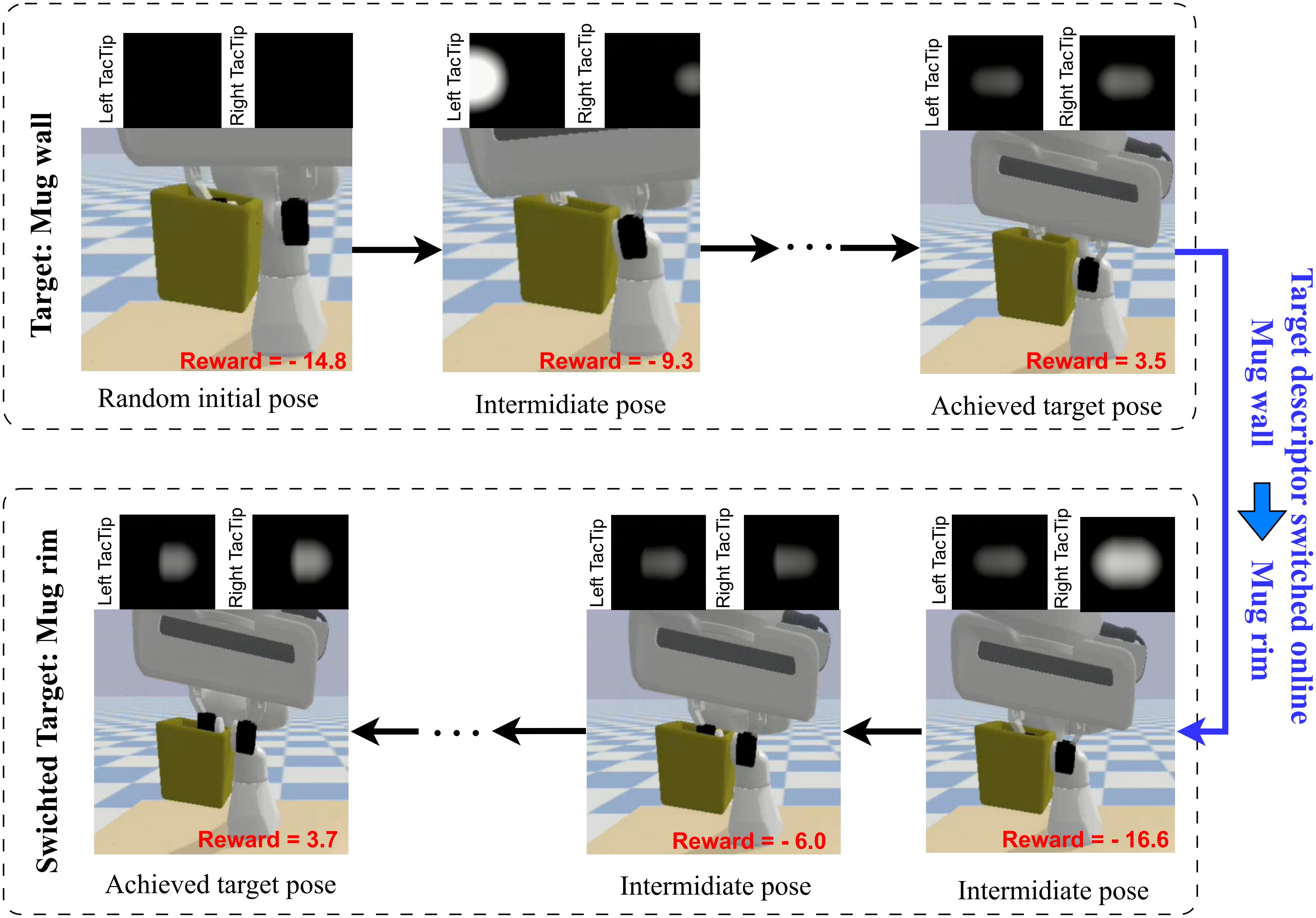}
  \vspace{-0.5em}
    \caption{An example of online adaptability of NeuralTouch RL policy with dynamic target changes demonstrated in a grasping task. 
    }
  \label{fig:online_change}
\end{figure}

Since we train our NeuralTouch to reach all target features conditioned on related neural descriptors corresponding to related target features, we are able to change the target of the policy in an online manner during test time (see Fig.~\ref{fig:online_change}). The vanilla RL-Touch policy cannot achieve this behaviour because it lacks the use of NDFs, further demonstrating the benefits of our proposed method.

\begin{figure}
  \centering
  \includegraphics[width=1\linewidth]{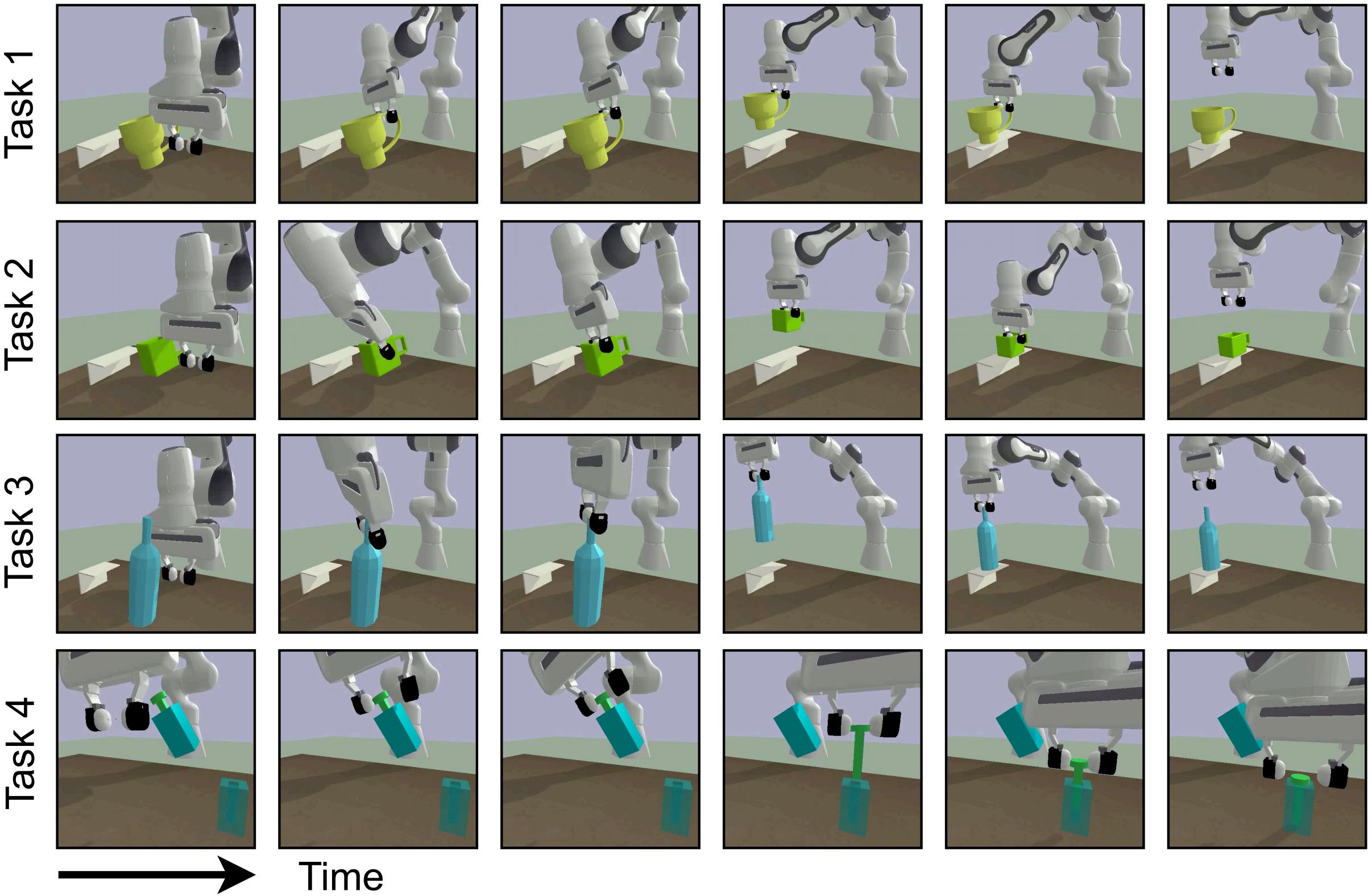}
    \caption{Snapshots of the robot performing four different tasks in simulation with NeuralTouch. From top to bottom row: object-pick-and-place (mug horizontal handle, mug rim, and bottle lid) and bolt-out/in-hole.}
  \label{fig:sim_exps}
\end{figure}

\subsection{Fine Manipulation Tasks in Simulation}
We further evaluated the performance of the above methods on several fine manipulation tasks in simulation, including object pick-and-place and bolt-out/in-hole tasks, with 60 trials conducted for each target feature (details in Sec.~\ref{subsec:task}). As shown in Table~\ref{table:sim_task}, our NeuralTouch method consistently outperforms all baselines across both tasks due to its high accuracy and robustness. Fig.~\ref{fig:sim_exps} presents representative execution snapshots for NeuralTouch.

For the bolt-out/in-hole task, C2FIL and C2FIL+RL-Touch achieve relatively high success rates, comparable to or exceeding NDF and NDF+RL-Touch. This behaviour is expected, as the bolt objects are highly similar in shape and geometry, which aligns well with the imitation nature of C2FIL. However, despite often reaching visually plausible grasp poses, NDF+RL-Touch and C2FIL+RL-Touch more frequently converge to slightly inaccurate grasping configurations, leading to insertion failures and thus lower success rates than NeuralTouch. This observation is consistent with the ablation results reported in Sec.~\ref{subsec:ablation}.

In the object pick-and-place tasks, the limitations of C2FIL-based methods become evident. Both C2FIL and C2FIL+RL-Touch perform much worse than NeuralTouch when manipulating objects with diverse target features, such as the mug rim and mug handle. Similar to NDFs+RL-Touch, these methods suffer from ambiguity in target feature identification, since different target features can induce similar contact geometries during interaction. Consequently, policies that rely on a single modality and demonstration-centric supervision struggle to generalise beyond the demonstrated feature instance.

By contrast, NeuralTouch explicitly integrates NDF-based visual guidance with tactile feedback during physical interaction, enabling the policy to disambiguate target features and execute precise manipulation actions. This joint multimodal reasoning leads to consistently high performance across all simulated manipulation tasks.


\begin{table}
\centering
\caption{PERFORMANCE ON SIMULATED MANIPULATION TASKS.\\TASK 1: PICK-AND-PLACE (MUG HORIZONTAL HANDLE).\\ TASK 2: PICK-AND-PLACE (MUG RIM). TASK 3: PICK-AND-PLACE (BOTTLE LID). TASK 4: BOLT OUT/IN HOLE.}
\addtolength{\tabcolsep}{-5pt}
\vspace{-0.5em}

\label{table:sim_task}
\begin{tblr}{
  vline{2-6} = {-}{},
  hline{1-5} = {-}{},
}
Tasks Index                            & NDFs    & NDFs+T  & C2FIL  & C2FIL+T & NT(Ours)        \\
1 & 40.0\% & 58.3\% & 46.7\% & 65.0\%  & \textbf{95.0\%} \\
2        & 56.7\% & 63.3\% & 45.0\% & 61.7\% & \textbf{96.7\%} \\
3     & 51.7\% & 76.7\% & 43.3\% & 56.7\%  & \textbf{93.3\%} \\
4                 & 11.7\% & 33.3\% & 71.7\% & 80.0\%  & \textbf{86.7\%} 
\end{tblr}
\vspace{-0.5em}

\end{table}

\subsection{Fine Manipulation Tasks in the Real World}\label{subsec:real_world_exp_results}
\begin{figure}%
    \centering
{\includegraphics[width=1.0\linewidth]{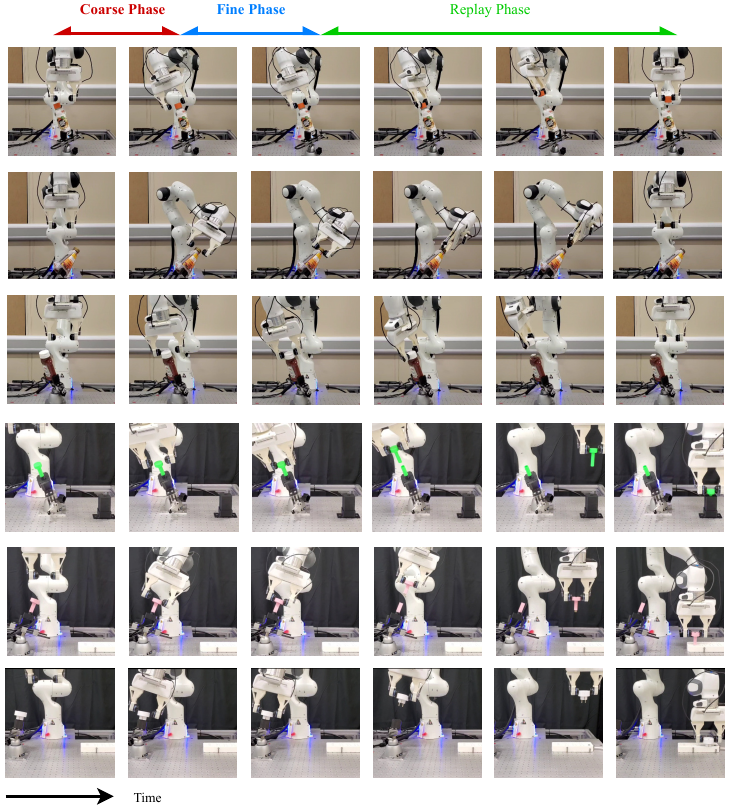} }%
    \caption{Robot arm equipped with a tactile gripper performing two real-world manipulation tasks requiring high accuracy. Top 3 rows: bottle-lid opening. 4th row: peg-in/out-hole insertion. Bottom 2 rows: to increase the difficulty of the insertion task, we also experimented with a USB-head bolt and a plug where the clearances were approximately $0.5\,\text{mm}$ and $1\,\text{mm}$, respectively.
    }%
  \label{fig:real_world_exp}
\end{figure}

We conducted sim-to-real NeuralTouch in two real-world fine manipulation tasks, including bottle-lid opening and bolt-out/in-hole tasks. The main pipeline is illustrated in Fig.~\ref{fig:framework} and the experimental setup is detailed in Sec.~\ref{subsec:task}. Again NeuralTouch consistently achieved much better performance in these tasks with all considered objects compared to the other four baselines (experimental results reported in Tables~\ref{table:bottle} and~\ref{table:bolt}). Some snapshots of the execution process when using NeuralTouch during the fine phase are shown in Fig.~\ref{fig:real_world_exp}. 
\subsubsection{Bottle-lid opening}
NeuralTouch achieves a 90\% success rate on the apple juice and ketchup bottles, and 85\% on the syrup bottle. In contrast, the vanilla NDF baseline only achieves success rates of 30\%-45\%, indicating that without tactile feedback, NDF-based policies frequently fail to open bottle lids. This is because the rotation action must be executed precisely around the central axis of the cylindrical lid; when the gripper approaches with a large positional offset (e.g., one finger significantly closer to the lid), the lid tends to oscillate forwards and backwards rather than rotate continuously. These failure modes are illustrated in the supplementary video. Adding tactile feedback to NDFs (denoted NDF+T) improves performance across all objects, increasing success rates to 65\%–80\%. This demonstrates that tactile sensing helps mitigate execution errors during lid rotation by providing local contact information. However, NDF+T still underperforms NeuralTouch, suggesting that simply augmenting NDFs with tactile input is insufficient to achieve high accuracy across varying lid geometries and surface properties.

The C2FIL and C2FIL+T baselines show a markedly different trend. Both methods perform well on the apple juice bottle (achieving 85\% and 90\% success, respectively), which is expected since this object is used during demonstration. However, their performance degrades substantially on the ketchup and syrup bottles, with success rates dropping to 10\%–40\%. This sharp decline indicates limited generalisability to objects with different shapes, textures, and frictional characteristics. Even with tactile feedback, C2FIL+T fails to transfer effectively beyond the demonstrated object, highlighting that imitation-based methods are prone to overfitting to demonstration-specific geometry and contact conditions.

Overall, these results show that while tactile sensing generally improves performance, NeuralTouch uniquely combines tactile feedback with a representation that generalises across object instances, enabling robust zero-shot sim-to-real bottle-lid opening beyond the demonstrated object.

\begin{figure}
  \centering
  \includegraphics[width=1\linewidth]{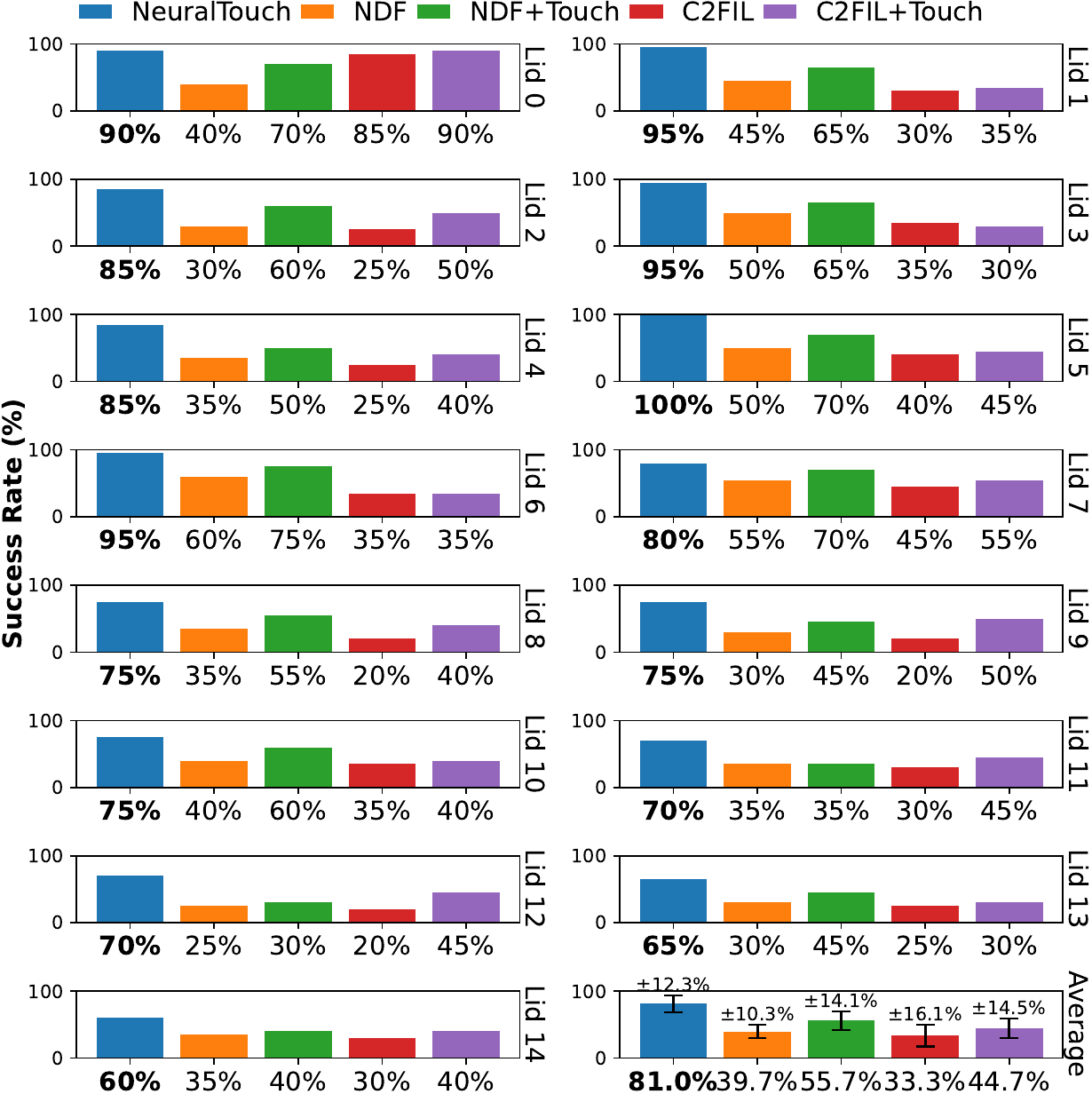}
    \caption{Success rates for the Bottle-Lid-Opening task across 15 different lids (the order matches the ones shown in Fig.~\ref{fig:bottle_lids}). NeuralTouch achieves the highest success rate on all evaluated bottle lids. 
    }
  \label{fig:baseline_bottle_lid_comparison}
\end{figure}

In addition, to further evaluate the generalisation capability of our method, we conducted bottle-lid-opening tasks using lids of various shapes and textures, as shown in Fig.~\ref{fig:bottle_lids}. The results are reported in Fig.~\ref{fig:baseline_bottle_lid_comparison}. Methods incorporating tactile feedback consistently outperform their non-tactile counterparts. Specifically, the average success rates across different lid variations are 33.3\% for C2FIL, 44.7\% for C2FIL+RL-Touch, 39.7\% for NDFs, and 55.7\% for NDFs+RL-Touch. Notably, C2FIL and C2FIL+RL-Touch perform well on the original demonstration lid (sample 0) but exhibit a substantial performance drop on other lids, indicating limited generalisability. In contrast, our NeuralTouch achieves an average success rate of 81.0\% across all lid variations, demonstrating strong robustness to shape and texture differences.

\subsubsection{Peg-out/in-hole}
Our NeuralTouch method achieved success rates of $55\%$, $25\%$ and $15\%$ for the bolt, plug and USB objects, respectively, consistent with the clearances of these objects progressively decreasing. Note that even though the success rates of NeuralTouch with the plug and the USB are not high, it does succeed sometimes. Also, even when the task fails on the insertion, it only has about 1\,mm error, compared to clearances of 1\,mm and 0.5\,mm respectively.

In contrast, the vanilla NDFs barely succeeds on the bolt (5\%) and consistently fails on the plug and USB insertion tasks. The primary cause of failure is misalignment of the grasping pose during insertion. C2FIL achieves a 25\% success rate on the bolt, which is used for demonstration, but also fails on the plug and USB tasks. C2FIL+RL-Touch improves the bolt success rate to 40\%, yet still fails on the plug and USB. The main limitation of C2FIL lies in its poor generalisability to unseen objects. Additionally, we observe that most failures with both the bolt and the USB result from misalignment between their principal axes and the grasping pose, leading to significantly increased frictional resistance that prevents successful withdrawal from the hole.

We also observe that the average performance on this task decreases from 86.7\% in simulation to 31.7\% in real-world experiments. We attribute this performance gap primarily to limitations in the current tactile real-to-sim transfer, which introduces discrepancies between simulated and real tactile observations. Addressing these limitations is beyond the scope of this paper; however, improving tactile real-to-sim transfer remains an important direction for future work.

Overall, the experiments show that NeuralTouch improves both grasping accuracy and generalisation by combining NDF-based visual guidance with tactile feedback. The advantage is most evident on unseen objects and ambiguous contact features, where single-modality baselines are more prone to failure. However, the real-world peg-out/in-hole results also reveal that the method remains limited by tactile sim-to-real transfer and by the sensitivity of high-precision insertion to small grasp misalignments.

\begin{table}
\caption{SIM-TO-REAL PERFORMANCE OF BOTTLE-LID-OPENING TASK USING DIFFERENT METHODS. 'T' STANDS FOR RL-TOUCH.}
\vspace{-0.5em}

\centering
\addtolength{\tabcolsep}{-5pt}
\begin{tabular}{>{\hspace{0pt}}m{0.248\linewidth}|>{\hspace{0pt}}m{0.102\linewidth}|>{\hspace{0pt}}m{0.152\linewidth}|>{\hspace{0pt}}m{0.129\linewidth}|>{\hspace{0pt}}m{0.179\linewidth}|>{\hspace{0pt}}m{0.1\linewidth}} 
\hline
success rate & NDFs & NDFs+T & C2FIL & C2FIL+T & NT (Ours)~ \\ 
\hline
apple juice & 40\% & 70\% & 85\% & \textbf{90\%} & \textbf{90\%} \\ 
\hline
ketchup & 45\% & 80\% & 20\% & 40\% & \textbf{90\%} \\ 
\hline
syrup & 30\% & 65\% & 10\% & 25\% & \textbf{85\%}
\end{tabular}
\vspace{-0.5em}

\label{table:bottle}
\end{table}
\begin{table}
\caption{SIM-TO-REAL PERFORMANCE OF PEG-OUT/IN-HOLE TASK USING DIFFERENT METHODS.}
\vspace{-0.5em}
\centering
\addtolength{\tabcolsep}{-5pt}
\begin{tabular}{>{\hspace{0pt}}m{0.248\linewidth}|>{\hspace{0pt}}m{0.102\linewidth}|>{\hspace{0pt}}m{0.152\linewidth}|>{\hspace{0pt}}m{0.129\linewidth}|>{\hspace{0pt}}m{0.179\linewidth}|>{\hspace{0pt}}m{0.1\linewidth}} 
\hline
success rate & NDFs & NDFs+T & C2FIL & C2FIL+T & NT (Ours)~ \\ 
\hline
bolt & 5\% & 30\% & 25\% & 40\% & \textbf{55\%} \\ 
\hline
plug & 0\% & 5\% & 0\% & 0\% & \textbf{25\%} \\ 
\hline
usb & 0\% & 0\% & 0\% & 0\% & \textbf{15\%}
\end{tabular}
\vspace{-0.5em}
\label{table:bolt}
\end{table}

\section{DISCUSSION AND FUTURE WORK} \label{sec:discussion}
\textbf{Summary.} We presented NeuralTouch, a new method to achieve accurate robotic grasping that integrates vision and touch to enable precise manipulation with various objects and target features of those objects. Our approach consists of two main phases: a coarse phase, where the NDF is used to generate an initial grasping pose; and a fine phase, where the robot engages in tactile servoing using a neural descriptor-based RL tactile policy upon approaching the initial pose. Additionally, we demonstrate applications of our method by introducing a third replay phase, where the robot performs downstream tasks requiring high precision, such as peg-out/in-hole. Our ablation study shows that NeuralTouch significantly outperforms baseline methods in grasping accuracy and generalizability. Furthermore, our method is sim-to-real transferable, which makes it easy to deploy in real-world scenarios. 

While works by Bauza et al.~\cite{bauza2024simple} and Zhao et al. \cite{zhao2023skill} are similar to ours in this regard, they differ in explicitly predicting only the in-hand contact pose for known object shapes; thus, they do not perform real-time closed-loop control and are unable to react to external disturbances or generalize to unseen objects. Our method implicitly learns to achieve the desired grasping pose for various unseen objects. 


\textbf{Limitations and future work.} While our NeuralTouch system can achieve accurate grasping poses, real-world applications could benefit from sub-millimetre precision. One limitation affecting precision is the accuracy of real-to-sim tactile transfer, particularly for very light contacts. Furthermore, our system currently lacks feedback control during the downstream task execution phase, where leveraging contact information could further enhance performance.\\
Ongoing studies \cite{zhao2024fots, xu2022efficient, higuera2023extrinsic, kim2022active, dong2021insertion} are complementary to our method, as they focus on learning physically interactive policies for insertion under the assumption that the object is already grasped in a suitable pose. These works highlight promising directions for future research, in which NeuralTouch could be integrated with such approaches to achieve more robust and generalisable manipulation.

\bibliographystyle{unsrt}
\bibliography{manual}
\end{document}